\documentclass[11pt,a4paper]{article}
\usepackage{times,latexsym}
\usepackage{url}
\usepackage[T1]{fontenc}

\usepackage{microtype}
\usepackage{booktabs}
\usepackage{multirow}
\usepackage{enumitem}
\usepackage{xurl}

\usepackage{graphicx}
\graphicspath{ {./images/} }

\usepackage{xspace}

\usepackage[acceptedWithA]{tacl2021v1}
\usepackage{xspace,mfirstuc,tabulary}

\usepackage{tablefootnote}

\newif\iftaclinstructions
\taclinstructionsfalse 
\iftaclinstructions
\renewcommand{\confidential}{}
\renewcommand{\anonsubtext}{(No author info supplied here, for consistency with
TACL-submission anonymization requirements)}
\newcommand{\instr}
\fi

\iftaclpubformat 

\else

\fi

\title{AfriSpeech-200: Pan-African Accented Speech Dataset for Clinical and General Domain ASR}

\author{\normalsize Tobi Olatunji$^{1*}$, Tejumade Afonja$^{2,3*}$, Aditya Yadavalli$^{4*}$, Chris Chinenye Emezue$^{5,6*}$ \\ \textbf{\normalsize Sahib Singh$^{7*}$, Bonaventure F.P. Dossou$^{5,6,8,9*}$, Joanne Osuchukwu$^{1}$,  Salomey Osei$^{10*}$, }\\
\textbf{\normalsize Atnafu Lambebo Tonja$^{8,11,12*}$, Naome Etori$^{13*}$, Clinton Mbataku$^{3*}$}\\ 
\\
\footnotesize
$^*$Masakhane NLP, $^1$Intron Health, $^2$CISPA Helmholtz Center for Information Security $^{3}$AI Saturdays Lagos $^4$Karya Inc,  \\
\footnotesize 
$^5$Mila Quebec AI Institute, 
$^6$Lanfrica, $^7$Ford Motor Company, $^8$Lelapa AI, $^9$McGill University, $^{10}$University of Deutso, \\
\footnotesize 
$^{11}$Instituto Politécnico Nacional, $^{12}$University of Colorado Colorado Springs, $^{13}$University of Minnesota\\
\texttt{Corresponding Author: tobi@intron.io}
}

\date{}

\begin{document}
\maketitle
\begin{abstract}
Africa has a very low doctor-to-patient ratio. At very busy clinics, doctors could see 30+ patients per day-- a heavy patient burden compared with developed countries-- but productivity tools such as clinical automatic speech recognition (ASR) are lacking for these overworked clinicians. However, clinical ASR is mature, even ubiquitous, in developed nations, and clinician-reported performance of commercial clinical ASR systems is generally satisfactory. Furthermore, the recent performance of general domain ASR is approaching human accuracy. However, several gaps exist. Several publications have highlighted racial bias with speech-to-text algorithms and performance on minority accents lags significantly. To our knowledge, there is no publicly available research or benchmark on accented African clinical ASR, and speech data is non-existent for the majority of African accents. We release AfriSpeech, 200hrs of Pan-African English speech, 67,577 clips from 2,463 unique speakers across 120 indigenous accents from 13 countries for clinical and general domain ASR, a benchmark test set, with publicly available pre-trained models with SOTA performance on the AfriSpeech benchmark.
\end{abstract}

\section{Introduction}

The African continent and the nearby islands constitute one-fourth of the land surface of the earth \cite{lodhi1993language}. Approximately 1.3 billion people live in Africa, which is about 18\% of the world's population \cite{enwiki:1132870977}. Of the estimated 7,000+ languages and dialects in the world, over 3,000 languages are found in Africa \cite{enwiki:1133594141, heine2000african}. 

Despite its large and predominantly young population, Africa bears a significant proportion of the global disease burden \cite{de2010tackling} with multiple socioeconomic factors contributing to high mortality and morbidity rates \cite{baingana2006changing}. Healthcare systems are overburdened and underfunded in many African countries \cite{oleribe2019identifying, naicker2009shortage,nkomazana2015stakeholders}, struggling to cope with the increasing demand for services, while at the same time facing significant shortages of trained health workers \cite{whoChronicStaff, ahmat2022health, naicker2010shortage, nkomazana2015stakeholders, kinfu2009health,etori2023we}. A recent study conducted by \citet{ahmat2022health} in 47 African countries shows that the region has a ratio of 1.55 health workers (physicians, nurses, and midwives) per 1000 people 3x less than the WHO-recommended density of 4.45 health workers per 1000 people. 

While technology can help mitigate some of these problems, \citet{bukachi2007information} and \citet{manyati2021systematic} aptly show that although Africa has enjoyed massive growth in mobile technology, telecommunication, and internet penetration over the past two decades, healthcare technology lags significantly. 

A 2019 systematic review on the use of Automatic Speech Recognition (ASR) for clinical documentation in the US from 1990 to 2018 by \citet{blackley2019speech} and other similar studies \citep{goss2019clinician, blackley2020physician, ahlgrim2016introduction, vogel2015analysis} showed that the use of speech recognition led to a 19-92\% decrease in mean documentation time, 50.3-100\% decrease in turnaround time, and 17\% improvement in documentation quality. 
However, in the African context, the lack of training datasets for many of the 3000+ languages and accents in the continent remains an obstacle in developing and adopting robust speech recognition systems for the general domain and for clinical ASR in particular \citep{doumbouya2021using, siminyu2021ai4d, babirye2022building, ogayo2022building}. While recent efforts have begun to turn this tide for the majority of African languages like Swahili, Kinyarwanda, and Yoruba \citep{gutkin2020developing, dossou2021okwugb, Olaleye2022YFACCAY}, over a thousand African languages and accents remain excluded from global speech research advancements. 

Recent single-digit word error rates (WER) \citep{chen2022wavlm, radford2022robust, hsu2021hubert, Baevski2020wav2vec2A} in multiple SOTA publications and benchmarks on Librispeech \citep{panayotov2015librispeech}, TED-LIUM3 \citep{hernandez2018ted}, and other datasets using architectures like Wav2vec2 \cite{Baevski2020wav2vec2A}, Conformer \cite{gulati2020conformer}, Transducer, and Whisper \cite{radford2022robust} contrast significantly with ASR performance for African accented speech \citep{gutkin2020developing, dossou2021okwugb} (see Figure \ref{appendix:wer_libri_afri}). We explore whether curating a large pan-African speech corpus might unlock comparable single-digit performance on African accents. We restrict this investigation to accented speech in English because English is the official language for the medical record in most Anglophone African countries, expanding the utility of this work to multiple Anglophone African countries. 


Our contributions are as follows: 
\begin{itemize}
\item We present \textit{AfriSpeech-200}\footnote{https://huggingface.co/datasets/tobiolatunji/afrispeech-200}, the first and most diverse open-source pan-African accented English speech corpus for clinical and general domain ASR, providing 200.70 hrs of accented speech, 67,577 speech-transcript pairs in 120 African accents across 13 countries, a benchmark dataset that paves the way for out-of-distribution, few-shot and zero-shot analyses on very-low-resource accents. \footnote{AfriSpeech-200 is licensed under a CC BY-NC-SA 4.0 license}


\item We present a templating framework to augment existing corpora with native African proper nouns and evaluate multiple SOTA pre-trained models and leading commercial ASR systems on our benchmark dataset. We provide in-depth analysis of selected models to explain their failure modes and offer helpful insights.

\item We fine-tune the best-performing open-source models and achieve SOTA performance on the AfriSpeech benchmark dataset (108 African accents) as well as show promising zero-shot performance on very low-resource accents. We provide best models\footnote{https://huggingface.co/Seyfelislem/afrispeech\_large\_A100} as publicly available pre-trained checkpoints.

\end{itemize}

\section{Related Work}


With the advent of large multilingual speech datasets \citep{7178964, indicwav2vec, Chen2021GigaSpeechAE, Ardila2020CommonVA, Valk2021VOXLINGUA107AD}, various research groups have proposed large self-supervised speech models such as wav2vec \cite{schneider19_interspeech}, vq-wav2vec \cite{Baevski2020vqwav2vecSL}, wav2vec 2.0 \cite{Baevski2020wav2vec2A}, HuBERT \cite{9585401}, XLSR \cite{Conneau2021UnsupervisedCR}, and XLS-R \cite{Babu2022XLSRSC}. These models achieved state-of-the-art performance on many downstream tasks such as automatic speech recognition (ASR), automatic speech translation (AST), and language identification. However, most existing systems still perform poorly on accented speech \citep{javed2022towards}. \citet{koenecke2020racial} further showed that popular commercial ASR systems -- like Amazon, Apple, Google, IBM, and Microsoft -- exhibit substantial racial disparities in their speech recognition capabilities. Most ASR systems work best for native English speakers and their accuracy plummets dramatically with non-native English speakers \cite{hassan2022improvement, prasad-jyothi-2020-accents}.

To enhance the performance of accented speech recognition, various methods have been proposed, which can be categorized into modeling and dataset approaches. On the modeling front, there have been efforts such as dialect-aware ASR models \cite{Yadavalli2022MultiTaskEM}, domain adversarial training (DAT) \cite{Sun2018DomainAT}, combining DAT with transfer learning \cite{Chen2020AipnetGA}, using voice conversion (VC) \cite{zhang22n_interspeech}, combining VC with speed perturbation \cite{zhang22n_interspeech}, and accent pre-training (Acc-PT) \cite{Das2021BestOB}. These efforts, however, produced marginal improvements and still exhibit poor generalization capabilities. 

Datasets have played a major role in improving ASR performance. The current SOTA in ASR \cite{radford2022robust} demonstrated the superior utility of large supervised datasets. 
Therefore, to bridge the ASR performance gap for African accented speech, multiple dataset creation efforts \citep{doumbouya2021using, siminyu2021ai4d, babirye2022building, ogayo2022building, gutkin2020developing, dossou2021okwugb, afonja2021learning, kamper2011multi,ibejih2022edustt} have been established. However, many of these datasets are limited in size and diversity. For example, Common Voice \citep{Ardila2020CommonVA} contains less than 10 hours of African English speech, \citet{li2021accent} evaluates on 50 hrs of African accented English (not released), \citet{sanabria2023edinburgh} provides 40 hrs of accented English, less than 20\% is African. \citet{kamper2011multi, de2007human} are limited to a few South African accents, and \citet{ibejih2022edustt} contains less than 8 hours, while \citet{afonja2021learning} contains less than 2 hours of accented African English speech. Furthermore, there are no available benchmarks for clinical ASR for African languages, creating a need for evaluation datasets that help identify areas of improvement in this domain. 

While previous works have primarily focused on adapting Western accents to African accents, to the best of our knowledge, there has been limited research specifically addressing domain adaptation from a general domain to the clinical domain in the African context. In this regard, our work is the first attempt to bridge this gap and tackle the unique challenges associated with adapting accented 
African English ASR systems to the clinical domain. 

\section{AfriSpeech Dataset}


We introduce AfriSpeech, a Pan-African accented English speech dataset for clinical and general domain ASR crowd-sourced from 2,463 African speakers, 200.70 hrs with an average audio duration of 10.7 seconds. Speaker, gender, age group, and clip domain distributions are shown in Table \ref{tab:dataset_stats}. In the following subsections, we describe the dataset creation process.

\subsection{Focus Languages}

We conducted an investigation on 120 African accents across 13 countries including the United States and Turkey. These accents originate from languages that belong to five language families, as documented by Eberhard \citep{Ethnologue_Eberhard}: Afro-Asiatic, Indo-European, Khoe-Kwadi (Hainum), Niger-Congo, and Nilo-Saharan. This selection represents the diverse linguistic landscape across western, eastern, and southern Africa. In Table \ref{tab:countries}, we provide an overview of the number of clips, speakers, and hours of data per country, with Nigerian accents comprising 67\% of the dataset. Since some languages are spoken across several countries (e.g., Swahili, isiZulu, Hausa, and Luganda), accents are not unique to countries.



\begin{table}
\small
\centering
\begin{tabular}{l|r|r|r}
\hline
\textbf{Country} & \textbf{Clips} & \textbf{Speakers} & \textbf{Hours}\\
\hline
Nigeria &  45875 &      1979 & 142.40 \\
Kenya &   8304 &       137 &  20.89 \\
South Africa&   7870 &       223 &  22.69 \\
Ghana &   2018 &        37 &   5.16 \\
Botswana &   1391 &        38 &   3.96 \\
Uganda &   1092 &        26 &   2.89 \\
Rwanda &    469 &         9 &   1.47 \\
United States\tablefootnote{Although the self-reported country from the speakers is the United States, their reported accents, namely Yoruba and Igbo, is mostly spoken in the western part of Africa.} &    219 &         5 &   0.53 \\
Turkey\tablefootnote{Even though the reported country is Turkey, the reported Zulu accent is mostly spoken in the southern part of Africa.}  &     66 &         1 &   0.18 \\
Zimbabwe &     63 &         3 &   0.18 \\
Malawi &     60 &         1 &   0.15 \\
Tanzania &     51 &         2 &   0.18 \\
Lesotho &      7 &         1 &   0.02 \\
\hline
\end{tabular}
\caption{Contributions by Country showing speakers, number of clips, and speech duration in seconds and hours.}
\label{tab:countries}
\end{table}

\subsection{Obtaining AfriSpeech Transcripts}

Neural network models learn concepts from training data. Where the training data is  predominantly Western (e.g. Common Voice \citep{ardila2019common}), the resulting ASR systems fail to capture important pan-African contexts. For example, ASR systems fail woefully at transcribing African names like "Ogochukwu" (Igbo), "Malaika" (Swahili), or "Uwimana" (Rwandan), while excellently transcribing Western names like "Lauren" and "Bryan"-- representative of the bias in their training corpora. 
To solve the problem of scarce African-centric text in the general and clinical domains, we created AfriSpeech using the following strategies.

\subsubsection{Finding Available Transcripts}
\label{findingtranscripts}
Our first task was to supplement existing large multi-domain corpora with African-centric text. Our first target was \textbf{Wikitext-103} \citep{merity2016pointer}, a collection of over 100 million tokens extracted from the set of verified "good" and "featured" articles on Wikipedia curated by Salesforce. We split this corpus on sentence boundaries and randomly sampled sentences for our transcript corpus. Our next strategy was \textbf{web scraping}. We crawled and scraped major African news websites across multiple African countries on topics like politics, entertainment, sports, religion, education, etc. In contrast to Wiki-text, the resulting corpus contained several African names, cities, and highly relevant vocabulary applicable to real-world use cases for downstream ASR. By scraping health-focused websites and health sections of news websites, we were able to get content from the clinical domain, albeit very little.

To increase clinical content representation, we focused on two multi-specialty biomedical datasets: \textbf{PubMed} \citep{wheeler2007database} and \textbf{NCBI disease} \citep{dougan2014ncbi}. We split these corpora on sentence boundaries and randomly sampled sentences for our transcript corpus. 

\subsubsection{Finding African Entities}
\label{finding_entities}
 We sourced for African-centric entities in two places: first, we leveraged an existing database of over 90,000 African names from the transatlantic slave trade between 1808 and 1863 \citep{anderson2013using}, which increased our coverage of African names, phonemes, and morphemes. We then used \citet{okagbue2017personal}'s dataset of 965 Igbo names collected to reflect the dialectal classification of Igbo people and supplemented it with 1,000 more Nigerian names from other cultures such as Yoruba, Hausa, Fulani, Tiv, Efik, Ibibio, etc. These names were obtained from freely available textbooks, online baby name websites, oral interviews, published articles, and online forums like Instagram and Twitter. Finally, we obtained a list of African cities from Wikipedia \citep{enwiki:1146587606}. 

\subsubsection{ AfriSpeech Templates}
 The web scraping corpus was highly relevant but small. In the larger biomedical and Wikitext datasets, African content was sparse. We, therefore, sought to increase the utility of the curated corpora by creating "Africanized" versions. Several studies have demonstrated the utility of "templates" as an effective way to create richer, more expressive training datasets, especially for Question-Answering and prompt engineering \citep{pawar2016question, brown2020language, yao2022prompt} and named entity recognition \citep{DBLP:conf/tsd/DavodyAKK22}. Inspired by this approach, we augment our dataset by sampling sentences from the corpora described above in addition to template sentences contributed by professional clinicians, hand-crafting a total of 140 template sentences. For each template sentence, we masked proper nouns (first names, last names, organizations, and cities), replacing them with their corresponding NER tags [PER, ORG, LOC]. We then randomly replaced the masked tokens with African-centric entities-- African names and cities, derived from section \ref{finding_entities} above, as well as common tropical diseases and medications. Each template sentence was reused 200 times. A random subset was sampled, sent as prompts for recording, and included with this release. Templated sentences represent approximately 30\% of this corpus.


\subsection{Audio Recording}

\begin{figure}[t]
\includegraphics[scale=0.2]{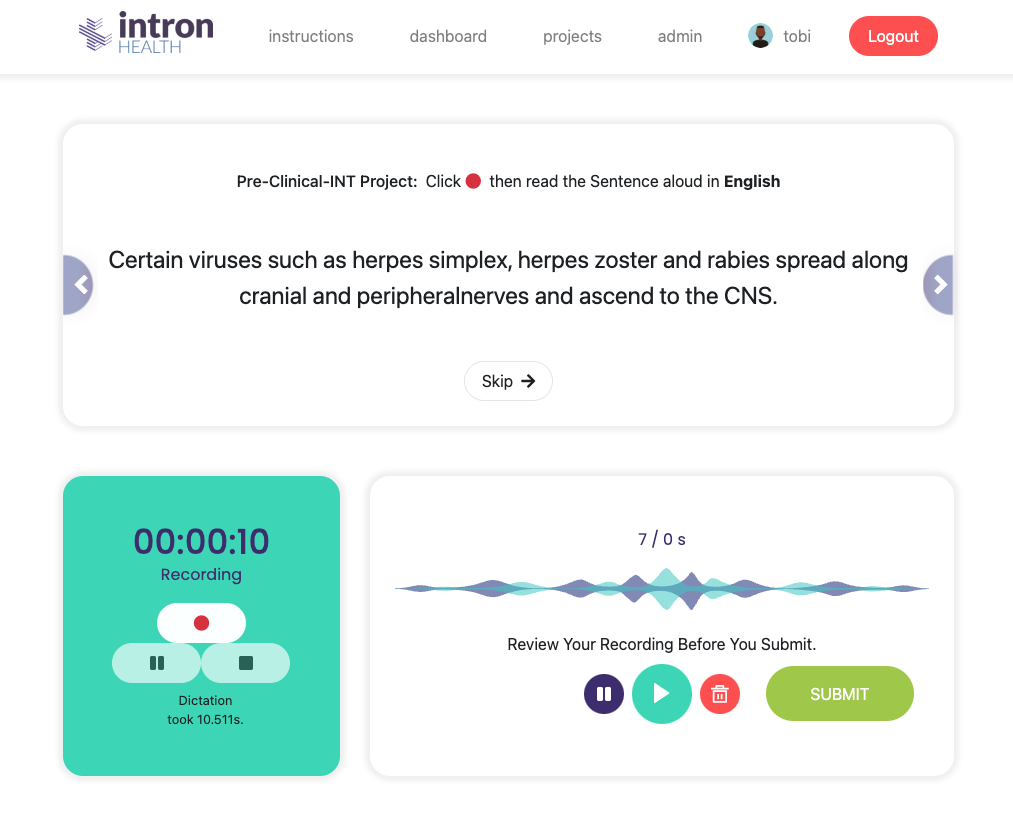} 
\centering
\caption{Intron Online Recording platform.}
\label{fig:web_app_screen}
\end{figure}

\paragraph{Collection:} Inspired by Common-Voice \citep{ardila2019common} and SautiDB \citep{afonja2021learning}, we developed and deployed a web-based application in Python/Flask (Figure \ref{fig:web_app_screen}) to collect crowd-sourced speech samples. The application also facilitates tracking of completion status, user demographics, reviews, and quality control.  The app presents randomly selected sentences (prompts) to the speakers and prompts them to record their voices while reading the text. The speech recordings are persisted as mono-channel, 16-bit wav files, with a 48 kHz sampling rate.  Post-processing tasks were performed on the audio recordings to remove samples shorter than 2 seconds and longer than 17 seconds. Raw unedited samples are provided as part of this release. Speakers in this dataset have been de-identified. Demographic information available includes gender, age group, accent, and country.

\paragraph{Annotation Instructions} Recorder demographics are presented in Table \ref{tab:dataset_stats}. Instructions were provided to crowd-sourced recorders as detailed in Appendix \ref{appendix:annotation}. Notably, the recorders were instructed to read punctuation marks in full and encouraged to use their natural accent.

\subsection{Quality Control}

  \paragraph{Projects:} Transcripts were bucketed into projects to separate clinical from general domain prompts. This approach maximized the time value of clinician contributors focusing their efforts more on medical prompts. 
  
 \paragraph{Reviewers:} We hired a team of human reviewers who up-voted or down-voted clips to indicate quality. Text feedback was also provided to recorders in 30\% of cases where negative feedback was indicated. The text feedback contained the reason for the down-vote and was intended to help recorders improve future recording quality. 
 
 \paragraph{Guest Clip Review:} New recorders were admitted as guests and allowed to record a maximum of 200 clips before quality review. 10 to 30 clips were reviewed per guest and those who passed review were promoted to a "Paid" status. 

 \paragraph{Paid Clip Review:} In the paid category, users were allowed a maximum of 200 clips before a temporary pause for quality check. During the temporary suspension, reviewers randomly reviewed 10\% of the speech samples provided and positive, negative, or text feedback was provided. Access was restored if quality remained satisfactory, or users were blacklisted if over 30\% of clips reviewed were down-voted.
 
 
 \paragraph{Delisting Problematic Sentences:}  Where an audio clip receives a down-vote, the corresponding sentence is released for re-recording by a different user. If a clip recorded for the same sentence receives a second down-vote, the transcript itself is blacklisted.

\section{Experiments}

\subsection{Data}

\begin{table}
\small
\centering
\begin{tabular}{|l|l|}
\hline
\multicolumn{2}{|l|}{\textbf{Speaker Gender Ratios - \# Clip \%}}\\
\hline
Female & 57.11\%  \\
Male & 42.41\%  \\
Other/Unknown & 0.48\% \\
\hline
\multicolumn{2}{|l|}{\textbf{Speaker Age Groups - \# Clips}}\\
\hline
<18yrs & 1,264 (1.87\%) \\
19-25 & 36,728 (54.35\%)  \\
26-40 & 18,366 (27.18\%)  \\
41-55 & 10,374 (15.35\%) \\
>56yrs & 563 (0.83\%)  \\
Unknown & 282 (0.42\%) \\
\hline
\multicolumn{2}{|l|}{\textbf{Clip  Domain - \# Clips}}\\
\hline
Clinical & 41,765 (61.80\%)  \\
General & 25,812 (38.20\%) \\
\hline
\end{tabular}
\caption{Dataset statistics.}
\label{tab:dataset_stats}
\end{table}

AfriSpeech-200 is a manually reviewed and curated subset, representing 7\% of the total AfriSpeech dataset, intended as an initial public release to stimulate research into African clinical and general domain ASR for accents with little or no representation in speech research. Table \ref{tab:countries} shows the distribution of clips, unique speakers, and hours by country. 

\begin{table}
\small
\centering
\begin{tabular}{l|c|c|c}
\hline
\textbf{Item} & \textbf{Train} & \textbf{Dev}  & \textbf{Test}\\
\hline

\# Speakers & 1466 & 247 & 750 \\
\#  Hours & 173.4 & 8.74 & 18.77 \\
\#  Accents & 71 & 45 & 108 \\
Avg secs/speaker & 425.80 & 127.32 & 90.08 \\
clips/speaker & 39.56 & 13.08 & 8.46 \\
speakers/accent & 20.65 & 5.49 & 6.94 \\
secs/accent & 8791.96 & 698.82 & 625.55 \\
\# general domain & 21682 & 1407 & 2723 \\
\# clinical domain & 36318 & 1824 & 3623 \\

\hline
\end{tabular}
\caption{Dataset splits showing speakers, number of clips, and speech duration in Train/Dev/Test splits.}
\label{tab:splits}
\end{table}

As shown in Table \ref{tab:splits}, the train, test, and development sets are bucketed such that any given speaker may appear in only one. This ensures that contributors seen at train time are not seen at test time, which would skew the results. 

\subsection{Benchmarks}

We compare SOTA open-source pre-trained ASR models: Whisper \citep{radford2022robust}, Wav2vec2 \citep{Baevski2020wav2vec2A}, XLSR \citep{Babu2022XLSRSC}, Hubert \citep{hsu2021hubert}, WavLM \citep{chen2022wavlm}, Conformer  \citep{gulati2020conformer}, and CRDNN-RNNLM \citep{ravanelli2021speechbrain} with commercial clinical and non-clinical ASR systems. We refer readers to read the respective papers for details on pretraining corpora, model architecture, and hyperparameters. For each model, we compare performance (WER) on Librispeech test-clean partition \citep{panayotov2015librispeech} with WER on the AfriSpeech dev and test sets. Single-run results are provided.

\begin{table*}
\tiny
\centering
\begin{tabular}{l|l|p{1.8cm}|p{0.7cm}|l|l|l|l|l|l}
\toprule 
Model & Params & Training/Fine-tuning Corpora & ls-clean & \multicolumn{3}{c|}{Dev (45 accents)} & \multicolumn{3}{c}{Test (108 accents)}\\
  & & &  & General & Clinical & Both & General & Clinical & Both \\
\midrule
\multicolumn{10}{l}{Open-Source SOTA Models}\\
\hline
openai/whisper-large     & 1550M & Multi, 680k hrs  &    0.167          &   0.235 &    0.287 &  0.261 &   0.240 &    0.375 &  0.306 \\
openai/whisper-medium     & 769M & Multi, 680k hrs &       0.166       &   0.246 &    0.300 &  0.273 &   0.276 &    0.392 &  0.332 \\
openai/whisper-medium-en   & 769M& Multi, 680k hrs   &     0.169          &   0.267 &    0.315 &  0.291 &   0.304 &    0.414 &  0.358 \\
openai/whisper-small    & 244M & Multi, 680k hrs  &          0.167           &   0.313 &    0.372 &  0.343 &   0.330 &    0.455 &  0.391 \\
openai/whisper-small-en  & 244M & Multi, 680k hrs   &      0.167  &   0.319 &    0.384 &  0.352 &   0.350 &   0.482 &  0.414 \\
nvidia/stt-en-conformer-ctc-large     & 118M & Multi, 10 &   0.210      &   0.410 &    0.486 &  0.448 &     - &      - &    - \\
nvidia/stt-en-conformer-transducer-large  & 139M & Multi, 10  &  0.150   &   0.408 &    0.477 &  0.443 &     - &      - &    - \\
jonatasgrosman/wav2vec2-large-xlsr-53-english  & 317M & Multi, 3  &  0.100  &   0.498 &    0.561 &  0.530 &   0.506 &    0.650 &  0.576 \\
jonatasgrosman/wav2vec2-xls-r-1b-english      & 317M & Multi, 4  &  0.087   &   0.502 &    0.571 &  0.537 &   0.521 &    0.670 &  0.594 \\

facebook/wav2vec2-large-960h-lv60-self   & 317M & Single, 2  &  0.051  &   0.512 &    0.587 &  0.550 &   0.533 &    0.694 &  0.611 \\
facebook/hubert-xlarge-ls960-ft      & 1B & Single, 1  &    0.052    &   0.531 &    0.610 &  0.571 &   0.562 &    0.725 &  0.641 \\
patrickvonplaten/wavlm-libri-clean-100h-large   & 317M & Single, 1   &  0.091  &   0.606 &    0.679 &  0.643 &   0.631 &    0.783 &  0.705 \\
facebook/wav2vec2-large-960h           & 317M & Single, 1   &   0.062   &   0.610 &    0.695 &  0.652 &   0.641 &    0.797 &  0.717 \\
facebook/wav2vec2-large-robust-ft-swbd-300h & 317M & Single, 5  & 0.093 &   0.689 &    0.778 &  0.734 &   0.733 &    0.906 &  0.817 \\
\hline
\multicolumn{10}{l}{Commercial ASR APIs}\\
\hline
Azure             & - & - &  &   0.438 &    0.468 &  0.453 &   0.340 &    0.444 &  0.391 \\
AWS         & - & - & &   0.332 &    0.437 &  0.385 &   0.354 &    0.536 &  0.442 \\
GCP        & - & - &  0.132  &   0.494 &    0.565 &  0.530 &   0.534 &    0.624 &  0.578 \\
\hline
\multicolumn{10}{l}{Commercial Clinical ASR APIs}\\
\hline
AWS [Medical] (Primary Care)        & -& -  & &   0.385 &    0.416 &  0.400 &   0.439 &    0.520 &  0.478 \\
GCP [Medical]        & - & - & &   0.550 &    0.475 &  0.512 &   0.567 &    0.537 &  0.552 \\
\hline
\multicolumn{10}{l}{Ours }\\
\hline
facebook/wav2vec2-large-xlsr-53-english-general & 317M & + AfriSpeech-general &  0.253 &   0.254 &    0.437 &  0.347 &   0.236 &    0.468 &  0.349 \\
facebook/wav2vec2-large-xlsr-53-english-clinical & 317M & + AfriSpeech-clinical  & 0.415 &   0.437 &    0.312 &  0.374 &   0.424 &    0.308 &  0.368 \\
facebook/wav2vec2-large-xlsr-53-english-all  & 317M & + AfriSpeech  &   0.314  &   0.295 &    0.308 &  0.302 &   0.279 &    0.308 &  0.293 \\
openai/whisper-medium-general       & 769M & + AfriSpeech-general  &  0.351          &   \textbf{0.205} &    0.486 &  0.347 &   \textbf{0.186} &    0.525 &  0.351 \\
openai/whisper-medium-clinical        & 769M & + AfriSpeech-clinical  &   0.568         &   0.491 &    0.264 &  0.376 &   0.464 &    0.266 &  0.368 \\
openai/whisper-medium-all       & 769M & + AfriSpeech  &       0.418           &   0.213 &    \textbf{0.241} &  \textbf{0.227} &   0.192 &    \textbf{0.242} &  \textbf{0.216} \\

\bottomrule
\end{tabular}
\caption{Results showing selected models, number of parameters, Number of pre-training/fine-tuning corpora ["Multi" refers to multilingual or multi-task], Librispeech \citep{panayotov2015librispeech} test clean WER and AfriSpeech dev and test set performance for open-source, commercial ASR models, and fine-tuned models (Ours). Missing values indicate incomplete or failed experiments.}
\label{tab:models_benchmarks}
\end{table*}

\subsection{Fine-tuning}
Based on the benchmark results in Table \ref{tab:models_benchmarks} and GPU memory constraints, 2 top performing open-source model architectures were selected for fine-tuning. Although commercial ASR systems outperformed many open-source models, they are excluded from fine-tuning experiments because their model architectures and underlying pre/post-processing logic are unknown. 

\paragraph{Selected Model Architectures}
\begin{enumerate}[leftmargin=0.4cm, noitemsep]
    \item wav2vec-large-xlsr-53 \citep{grosman2021xlsr53-large-english}: an Encoder-decoder architecture with CNN-based feature extractor, code book, and transformer-based encoder, 378.9M parameters; LR 1e-4.
    \item whisper-medium \citep{radford2022robust}: a Decoder-only multi-task architecture, 789.9m parameters; LR 2.5e-4.
\end{enumerate}

For each model, we fine-tuned with FP16, AdamW \citep{loshchilov2017decoupled}, batch size of 16, for 10 epochs, with a linear learning rate decay to zero after a warmup over
the first 10\% of iterations. We fine-tune and evaluate on 3 domains: (1) \textbf{general} (25,812 clips), (2) \textbf{clinical} (41,765 clips), and (3) \textbf{both} (67,577 clips). We train on each domain and test across all 3 domains to investigate the effect of out-of-domain data on model performance. XLSR models were trained on a single Tesla T4 GPU with 16GB GPU memory while Whisper and Conformer models were trained on RTX8000 GPU with 48GB GPU memory. Fine-tuning took 24-48 hrs for all domains.

\subsection{Model Vocabulary}
\label{model_vocabulary}
 Most pre-trained models define a limited vocabulary of only Latin alphabets with no numbers or punctuations \citep{Baevski2020wav2vec2A}. In stark contrast, numbers are critical in healthcare, e.g. blood pressure 130/80mmHg, or Lab results 0.428 mmol/L. Eliminating all numerical references in clinical text is dangerous and counterproductive. Post-processing to convert all numerical values to long form is imperfect so we retain numbers in their original form. For fine-tuning experiments, we define an alphanumeric vocabulary with semantically important punctuations, characters, and symbols commonly used in medical practice (colon, question mark, plus, etc). 

\subsection{Evaluation}
We report our results as WER on AfriSpeech dev and test sets in addition to domain and accent-specific performance. Results are compared with Librispeech \citep{panayotov2015librispeech} test set performance. We also report the zero-shot performance of fine-tuned models on unseen accents in the test set.

\section{Results and Discussion}

\subsection{Africa-centric Fine-tuning Improves Robustness} 
As shown in Table \ref{tab:models_benchmarks}, compared with its pre-trained version, xlsr-53 fine-tuned on general domain speech (AfriSpeech-general) yields 53.4\% relative improvement. Xlsr-53 fine-tuned on clinical domain speech (AfriSpeech-clinical) yields 52.6\%, and xlsr-53 fine-tuned on the combined domains (AfriSpeech-all) yields 49.1\% relative improvement. The trend is similar with pre-trained Whisper-medium, yielding 32.6\% relative improvement on the general domain, 32.1\% on the clinical domain, and 34.9\% when finetuned on combined domains. 

\begin{table}
\tiny
\centering
\begin{tabular}{l|l|l|l|l|l|l}
\toprule 
Accent & Samples & \multicolumn{1}{c|}{OpnSrc} & \multicolumn{3}{c|}{Commercial} & \multicolumn{1}{c}{Ours} \\
   &  & Whisper & Azure & GCP & AWS & Whisper \\
\midrule
\multicolumn{7}{l}{Niger-Congo}\\
\hline
Ukwuani &      119 &                                                                                       0.364 &  0.393 &  0.677 &  0.484 &                                                                                            \textbf{0.244} \\
Eggon &      100 &                                                                                          0.254 &  0.316 &  0.616 &  0.359 &                                                                                            \textbf{0.122} \\
Bini          &     76 &                                                                                         0.830 &  0.840 &  0.916 &  1.061 &                                                                                           \textbf{0.412} \\
Yoruba, hausa &      75 &                                                                                       0.462 &  0.367 &  0.463 &  0.437 &                                                                                         \textbf{0.133} \\

Ekpeye        &    70 &                                                                  0.376 &  0.406 &  0.582 &  0.539 &                                                                                          \textbf{0.190} \\
Bajju         &    61 &                                                                                       0.229 &  0.323 &  0.428 &  0.378 &                                                                                        \textbf{0.171} \\
Ikulu         &    60 &                                                                                      0.406 &  0.388 &  0.650 &  0.543 &                                                                                        \textbf{0.195} \\
Jaba          &    59 &                                                                                   0.462 &  0.475 &  0.798 &  0.529 &                                                                                         \textbf{0.268} \\
Ekene         &    55 &                                                                                       0.414 &  0.350 &  0.673 &  0.519 &                                                                                         \textbf{0.192} \\
Agatu         &    54 &                                                                              0.734 &  0.725 &  0.903 &  0.793 &                                                                                      \textbf{0.387} \\
Ijaw(nembe)     &    49 &                                                                                       0.478 &  0.529 &  0.743 &  0.675 &                                                                                         \textbf{0.275} \\

Delta         &    48 &                                                                               0.384 &  0.351 &  0.724 &  0.473 &                                                                                       \textbf{0.205} \\
Igarra        &    45 &                                                                                        0.591 &  0.539 &  0.839 &  0.687 &                                                                                         \textbf{0.258} \\
Khana         &     45 &                                                                                        0.539 &  0.584 &  0.761 &  0.785 &                                                                                         \textbf{0.318} \\
Gbagyi        &    42 &                                                                                     0.327 &  0.461 &  0.633 &  0.475 &                                                                                        \textbf{0.195} \\
Jukun         &    42 &                                                                                    0.182 &  0.234 &  0.415 &  0.244 &                                                                                        \textbf{0.122} \\
Brass         &    39 &                                                                                        0.147 &  0.269 &  0.357 &  0.309 &                                                                                        \textbf{0.131} \\                           
\hline
\multicolumn{7}{l}{Afro-Asiatic}\\
\hline
Mada          &    78 &                                                                                     0.485 &  0.560 &  0.684 &  0.634 &                                                                                         \textbf{0.236} \\
Mwaghavul     &     67 &                                                                                      0.444 &  0.513 &  0.690 &  0.613 &                                                                                         \textbf{0.235} \\
Angas         &    58 &                                                                                       0.605 &  0.580 &  0.862 &  0.653 &                                                                                         \textbf{0.343} \\
\bottomrule
\end{tabular}
\caption{Zero shot (OOD) accents. Test set WER on top 20 accents absent from the training set for open-source (OpnSrc), commercial, and fine-tuned ASR models (Ours).}
\label{tab:OOD_accents_WER}
\end{table}

\subsection{Training Data Bias} In the Open-Source section of Table \ref{tab:models_benchmarks}, AfriSpeech dev and test set performance correlates with the number and diversity of pre-training datasets. For example, Wav2vec2 models trained exclusively on Librispeech significantly underperform when compared with those trained on multiple \citep{Baevski2020wav2vec2A} or multilingual corpora \citep{Babu2022XLSRSC}. Models trained on Multilingual or multi-task corpora \citep{radford2022robust, gulati2020conformer} learn more useful representations, are more linguistically diverse, are more robust, and generalize better to accented speech. 

\begin{table*}
\tiny
\centering
\begin{tabular}{l|l|c|c|l|l|l|l|l|l|l|}
\toprule 
Accent & Country & Test Samples & Train Samples & \multicolumn{2}{c|}{\textbf{Open Source}} & \multicolumn{3}{c|}{\textbf{Commercial}} & \multicolumn{2}{c|}{\textbf{Ours, Finetuned}} \\
  &  &  & & xlsr-53 &  whisper & Azure & GCP & AWS & XLSR  & Whisper \\
\midrule
\multicolumn{10}{l}{Niger-Congo}\\
\hline
Yoruba & [NG] & 575 &14233& 0.576 &  0.327 & 0.364 & 0.581 & 0.421 & 0.291 &  \textbf{0.218} \\
Swahili & [KE, TZ, UG, ZA] & 485 &5484 &  0.448 & 0.192 & 0.307 &  0.436 &  0.305 & 0.244 & \textbf{0.181} \\
Igbo & [NG] & 319 &8068& 0.564 & 0.338 & 0.393 &  0.563 &  0.441 & 0.273 & \textbf{0.197} \\
Zulu &  [TR, LS, ZA] & 156 & 1309& 0.471 & \textbf{0.223} & 0.329 &  0.477 &  0.345 &   0.315 &   0.237 \\
Setswana &  [BW, ZA] & 96 &1275&  0.448 &  \textbf{0.208} & 0.288 &  0.446 &  0.300 & 0.291 &  0.234 \\
Isizulu  &  [ZA] & 88 & 779 &    0.457 & \textbf{0.182} &  0.254 &  0.406 &  0.292 & 0.265 &  0.206 \\
Ijaw  &  [NG] & 77 & 2371 &  0.608 & 0.364 & 0.372 &  0.671 &  0.446 & 0.321 &\textbf{0.238} \\
Luhya  &  [KE] & 69 & 426 &   0.538 &0.310 & 0.548 &  0.489 &  0.427 & 0.296 & 0.245 \\
Twi  & [GH] &  54 & 1321 &  0.504 & 0.184 & 0.382 &  0.510 &  0.361 & 0.236 & 
 \textbf{0.177} \\
Idoma  &  [NG] &  53 & 1767 & 0.607 &0.384 & 0.424 &  0.639 &  0.543 & 0.294 &  \textbf{0.243} \\
Luganda &  [KE, UG, BW] & 44 & 529 & 0.525 &0.320 & 0.362 &  0.526 &  0.378 &  0.381 & \textbf{0.277} \\
Tswana       &  [BW, ZA] & 34 & 289 &  0.362 & \textbf{0.184} &   0.265 &  0.425 &  0.267 & 0.249 &  0.241 \\
Akan (fante) & [GH] &  29 & 230 &   0.732  &   0.418 & 0.425 &  0.803 &  0.604 & 0.290  &  \textbf{0.197} \\
Kikuyu &      [KE] &       24 &  163&  0.406 &     0.160 &  0.275 &  0.387 &  0.300 &   0.221 &  \textbf{0.126} \\
Xhosa  &  [ZA] &  17 & 342&  0.498 &  0.265 &  0.322 &  0.332 &  0.389 &  0.318 & \textbf{0.237} \\
Sepedi  &   [ZA] & 17 &   176 &  0.651  &  0.373 &   0.394 &  0.659 &  0.458 & 0.414 & \textbf{0.285} \\
Kiswahili  &  [KE] &  16 & 811 &   0.466 &     \textbf{0.159} &  0.389 &  0.394 &  0.274 & 0.173 &0.163 \\
Urhobo &    [NG] &   15 &578&    0.551 &  0.378 &    0.423 &  0.678 &  0.423 &  0.345 &    \textbf{0.210} \\
Nembe & [NG] &  14 & 546 &  0.571 &   0.352 &  0.449 &  0.556 &  0.449 &  0.372 & \textbf{0.296} \\
Kinyarwanda  &  [RW] &  14 &439 &  0.495 & \textbf{0.216} &  0.338 &  0.527 &  0.437 &  0.369 &0.311 \\

\hline
\multicolumn{10}{l}{Afro-Asiatic}\\
\hline
Hausa & [NG] & 168 &5453& 0.627 & 0.358 & 0.457 &  0.633 &  0.488 & 0.320 & \textbf{0.243} \\

\hline
\multicolumn{10}{l}{Indo-European}\\
\hline
Afrikaans    & [ZA] & 49 &1911& 0.373 & \textbf{0.142} & 0.202 &  0.443 &  0.209 & 0.283 & 0.211 \\
\hline
\multicolumn{10}{l}{Nilo-Saharan}\\
\hline
Luo &  [UG, KE] & 12 &    179& 0.411 & 0.234 &  \textbf{0.229} &  0.343 &  0.343 & 0.309 & 0.234 \\
\bottomrule
\end{tabular}
\caption{Test set performance per accent  for open-source, commercial, and fine-tuned ASR models.}
\label{tab:top_accents_WER}
\end{table*}

\subsection{Clinical ASR is Sensitive to Model Vocabulary}
As mentioned in Section \ref{model_vocabulary}, most ASR models tend to transcribe numbers in their extended forms, which have a detrimental effect on their WER as shown in Table~\ref{tab:models_benchmarks}, particularly in the clinical domain where numerical values need to be transcribed accurately (column 6 \& 9). However, ASR models with a larger vocabulary, such as Whisper, Commercial ASR models, and our fine-tuned models, demonstrate superior performance by effectively transcribing numbers in clinical speech and converting them into correct numeric representations.

\subsection{Punctuation Prediction is Critical for Clinically Useful ASR} 
Medical documents typically follow preset sequence and formatting, for example, patient history, general examination, laboratory investigation, etc., separated by new lines, section titles, or semi-colons. Punctuation commands such as "Next line", "full stop" (.), "query" (?), "comma" (,), "colon" (:) are frequently used in healthcare dictations to add structure to documents. ASR systems without support for such commands force clinicians to review every line of the ASR transcript to add/revise punctuations and document structure, prolonging documentation time and patient wait time \citep{Sunkara2020}. As a result, commercial clinical ASR systems supporting these commands are preferable and outperform general-purpose models.


\subsection{Commercial ASR APIs are Not So Global} 
The 3 large commercial ASR systems evaluated in this study have global presence. Millions of African Android users have access to Voice typing through the Google keyboard and Microsoft Word users have access to its ASR engine. Table \ref{tab:top_accents_WER} compares the performance of these ASR APIs on majority African accents and we show that despite their global presence, performance lags significantly on some of Africa's most populous accents like Swahili and Yoruba. 

\subsection{Domain Adaptation} 
Pre-trained whisper models performed better on general domain speech (AfriSpeech-general) when compared with the clinical domain, demonstrating the relative domain-driven difference in difficulty despite the robust training data for Whisper models (680k hours, 90 languages). Cross-domain fine-tuning yields significant gains helping to somewhat bridge this gap. Our results agree with prior work on domain adaptation \citep{sun2017unsupervised, abdelwahab2015supervised} showing that models trained exclusively on clinical data improve when general domain data is added. Whisper shows 9\% relative improvement on the clinical domain with the addition of general domain data.  However, this trend is reversed with general domain data. Adding speech from the clinical domain leads to a 3\% and 18.2\% relative drop for Whisper and xlsr-53 respectively. Domain adaptation is no silver bullet. Care must be taken to apply this approach where benefits outweigh risks.



\subsection{Accent-level Performance} 
\label{accent_level_performance}
Table \ref{tab:top_accents_WER} shows test set performance on the top 23 AfriSpeech accents grouped by their language families. We report the results for open-source, commercial, and fine-tuned ASR models. Fine-tuned models (ours) average relative improvement is 26.7\% over the open-source ASR models and 36.5\% over the commercial ASR models. For several accents, we observe that the whisper model fine-tuned with our AfriSpeech dataset shows the best overall performance with an average relative improvement of 16.2\% across all accents, except in 4 South African languages (Zulu, isiZulu\footnote{We note that both Zulu and isiZulu are the same but they are labeled differently in our dataset. We further discuss this in the limitations section.}, Tswana, Afrikaans), Luo, and Kinyarwanda, where the fine-tuned model under-performs compared to the pretrained whisper model and commercial Azure model performs best on Luo accent. Although counter-intuitive, it is possible these accents are highly represented in Whisper pre-training data and require further investigation. 

\subsection{Zero-Shot Performance}
We further explore generalizability to unseen accents, i.e., out-of-distribution (OOD) accents. Table~\ref{tab:OOD_accents_WER} shows the results for the top 20 OOD accents in the test set. We observe an impressive 44.4\% relative performance improvement across all OOD accents with our fine-tuned Whisper model compared to the baselines and 49.8\% average relative improvement over the commercial models (Azure, GCP, AWS). These results demonstrate significant generalizability gains are achievable with better training data diversity.



\subsection{Take SOTA LibriSpeech Results with a Grain of Salt}  
\begin{figure*}
\includegraphics[scale=0.35]{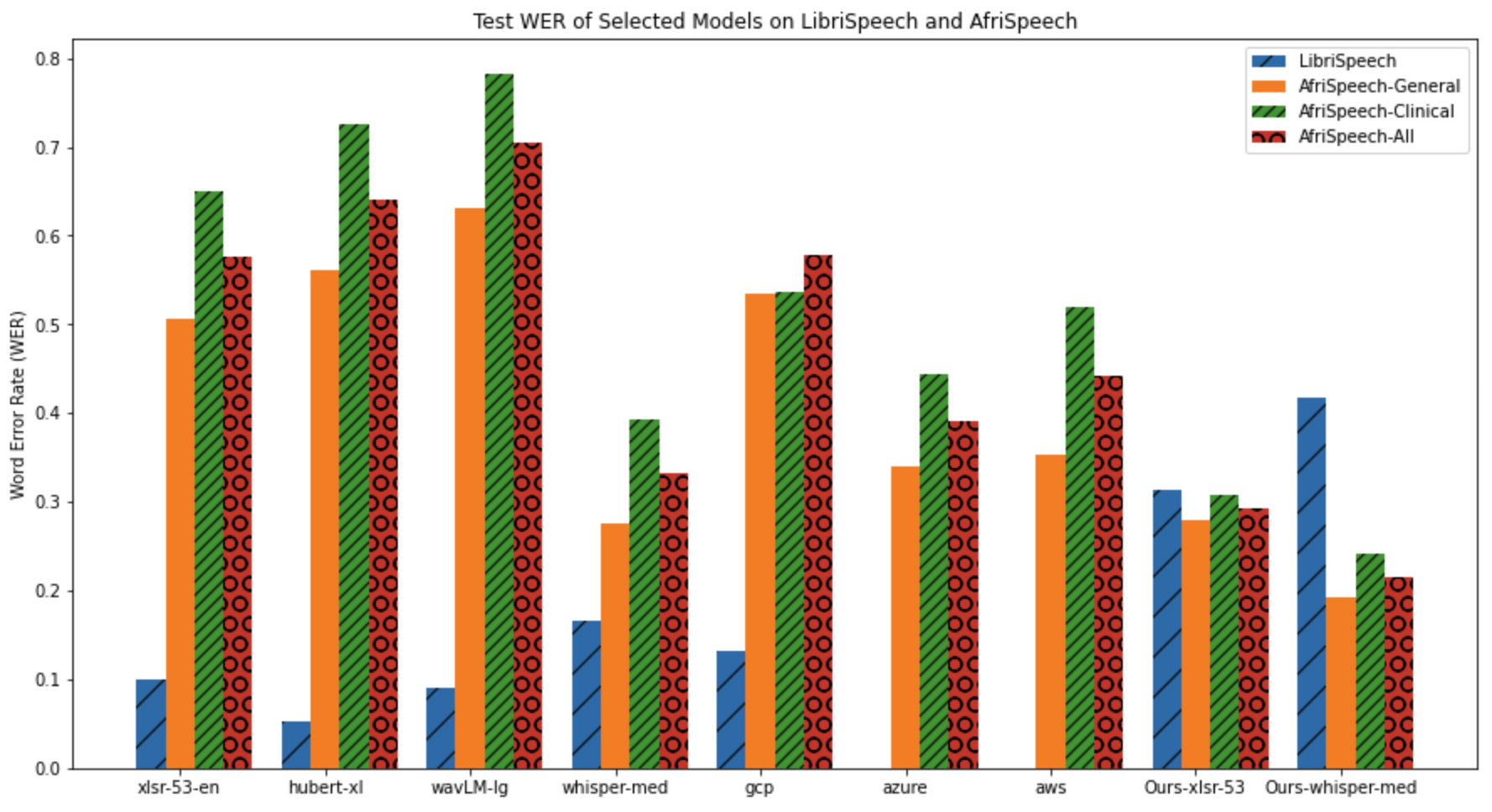} 
\centering
\caption{WER on LibriSpeech vs AfriSpeech for selected pre-trained models and commercial ASR systems.}
\label{appendix:wer_libri_afri}
\end{figure*}

Figure \ref{appendix:wer_libri_afri} contrasts LibriSpeech and AfriSpeech WER for several models. Many ASR leaderboards rank ASR models based on single-digit LibriSpeech \citep{panayotov2015librispeech} WER. Pre-trained ASR models, therefore, overfit to LibriSpeech at the expense of robust ASR performance for all people. As seen in Table \ref{tab:models_benchmarks}, several models are 3-10x worse on African accented speech with the exception of multi-lingual or multi-task models like Whisper, Conformer, and XLSR. 

\section{Limitations and Future Work}

\paragraph{Limited clinical Subdomains:} Although this dataset includes a variety of clinical text, several specialties are not represented. As a result, ASR performance may vary between clinical specialties.

\paragraph{Read Speech:} All audio samples in this release are read based on text prompts. Without appropriate augmentation, ASR Models trained on this dataset may underperform with conversational or spontaneous speech. 

\paragraph{North-African Accents} are not included in this work. Because of the distinct nature of those accents, performance on sub-Saharan accents may not necessarily generalize to the Northern African Region.

\paragraph{Self-reported Accents:} Similar to Common-Voice, recorders self-report their native tongue in free-text making it difficult to map to ISO-3 in all cases. Some users also reported their accents as "French", "English", "South African English", or a combination of accents. Although we attempted to clean and normalize the self-reported languages, this process was by no means perfect. As a result, accent names sometimes overlap e.g. Zulu and IsiZulu. Further cleanup could be done to consolidate these closely related accents. The dataset release will therefore include a normalized accent field for each sample. 

\paragraph{Medical Abbreviations are Inconsistent:} Since crowd-sourced recorders had varying levels of familiarity with the prompts, abbreviations like "Breast CA" may be pronounced fully as "Breast Cancer" or "Breast see-A". Since abbreviations abound in medical text and WER is not robust to such idiosyncrasies, models with correct predictions, e.g. "Breast Cancer" are sometimes wrongly penalized where the transcript reads "Breast CA".

\paragraph{Integrating ASR in Healthcare Settings is Challenging:}
 Cloud-based ASR presents some well-known challenges in healthcare. Privacy is a major concern as there is a risk of unauthorized or malicious third-party access to confidential patient information. Furthermore, the perceived higher value of healthcare data among malefactors also heightens security risks for hospitals and ASR vendors. Additionally, Unethical ASR vendors could misuse confidential data for model training and development without proper consent. 

\section{Ethical Considerations}

Clinical ASR models can improve productivity for clinicians, they can also increase documentation errors, especially through incorrect transcription of numbers, fractions, dates, and proper nouns which have legal, safety, and prognostic implications in healthcare. 
We caution clinicians to use ASR with full discretion and review transcripts carefully before final submission into the medical record. We release AfriSpeech hoping that it will be beneficial to clinical and non-clinical use cases within and outside Africa, improving ASR performance for accented speech and it may contain biases due to publicly available datasets. 
We do not have access to reviewers who are native speakers of most of the languages covered in AfriSpeech who can provide a rigorous review of self-reported accents. This hinders our ability to investigate samples from all languages. We hope that future users of the dataset will further investigate AfriSpeech's utility and quality for their languages.

\section*{Acknowledgments}
Tobi Olatunji acknowledges Intron Health for providing the dataset and compute resources. Chris Chinenye Emezue acknowledges the support of the Mila - Quebec AI Institute for compute resources.

\bibliography{tacl2021}

\begin{thebibliography}{78}
\expandafter\ifx\csname natexlab\endcsname\relax\def\natexlab#1{#1}\fi

\bibitem[{who()}]{whoChronicStaff}

\newblock {C}hronic staff shortfalls stifle {A}frica’s health systems: {W}{H}{O} study --- afro.who.int.
\newblock \url{https://www.afro.who.int/news/chronic-staff-shortfalls-stifle-africas-health-systems-who-study}.
\newblock [Accessed 15-Oct-2022].

\bibitem[{Abdelwahab and Busso(2015)}]{abdelwahab2015supervised}
Mohammed Abdelwahab and Carlos Busso. 2015.
\newblock Supervised domain adaptation for emotion recognition from speech.
\newblock In \emph{2015 IEEE International Conference on Acoustics, Speech and Signal Processing (ICASSP)}, pages 5058--5062. IEEE.

\bibitem[{Afonja et~al.(2021)Afonja, Mudele, Orife, Dukor, Francis, Goodness, Azeez, Malomo, and Mbataku}]{afonja2021learning}
Tejumade Afonja, Oladimeji Mudele, Iroro Orife, Kenechi Dukor, Lawrence Francis, Duru Goodness, Oluwafemi Azeez, Ademola Malomo, and Clinton Mbataku. 2021.
\newblock Learning nigerian accent embeddings from speech: preliminary results based on sautidb-naija corpus.
\newblock \emph{arXiv preprint arXiv:2112.06199}.

\bibitem[{Ahlgrim et~al.(2016)Ahlgrim, Maenner, and Baumstark}]{ahlgrim2016introduction}
Christoph Ahlgrim, Oliver Maenner, and Manfred~W Baumstark. 2016.
\newblock Introduction of digital speech recognition in a specialised outpatient department: a case study.
\newblock \emph{BMC medical informatics and decision making}, 16(1):1--8.

\bibitem[{Ahmat et~al.(2022)Ahmat, Okoroafor, Kazanga, Asamani, Millogo, Illou, Mwinga, and Nyoni}]{ahmat2022health}
Adam Ahmat, Sunny~C Okoroafor, Isabel Kazanga, James~Avoka Asamani, Jean Jacques~Salvador Millogo, Mourtala Mahaman~Abdou Illou, Kasonde Mwinga, and Jennifer Nyoni. 2022.
\newblock The health workforce status in the who african region: findings of a cross-sectional study.
\newblock \emph{BMJ Global Health}, 7(Suppl 1):e008317.

\bibitem[{Anderson et~al.(2013)Anderson, Borucki, Da~Silva, Eltis, Lachance, Misevich, and Ojo}]{anderson2013using}
Richard Anderson, Alex Borucki, Daniel~Domingues Da~Silva, David Eltis, Paul Lachance, Philip Misevich, and Olatunji Ojo. 2013.
\newblock Using african names to identify the origins of captives in the transatlantic slave trade: crowd-sourcing and the registers of liberated africans, 1808--1862.
\newblock \emph{History in Africa}, 40(1):165--191.

\bibitem[{Ardila et~al.(2019)Ardila, Branson, Davis, Henretty, Kohler, Meyer, Morais, Saunders, Tyers, and Weber}]{ardila2019common}
Rosana Ardila, Megan Branson, Kelly Davis, Michael Henretty, Michael Kohler, Josh Meyer, Reuben Morais, Lindsay Saunders, Francis~M Tyers, and Gregor Weber. 2019.
\newblock Common voice: A massively-multilingual speech corpus.
\newblock \emph{arXiv preprint arXiv:1912.06670}.

\bibitem[{Ardila et~al.(2020)Ardila, Branson, Davis, Henretty, Kohler, Meyer, Morais, Saunders, Tyers, and Weber}]{Ardila2020CommonVA}
Rosana Ardila, Megan Branson, Kelly Davis, Michael Henretty, Michael Kohler, Josh Meyer, Reuben Morais, Lindsay Saunders, Francis~M. Tyers, and Gregor Weber. 2020.
\newblock Common voice: A massively-multilingual speech corpus.
\newblock In \emph{LREC}.

\bibitem[{Babirye et~al.(2022)Babirye, Nakatumba-Nabende, Katumba, Ogwang, Francis, Mukiibi, Ssentanda, Wanzare, and David}]{babirye2022building}
Claire Babirye, Joyce Nakatumba-Nabende, Andrew Katumba, Ronald Ogwang, Jeremy~Tusubira Francis, Jonathan Mukiibi, Medadi Ssentanda, Lilian~D Wanzare, and Davis David. 2022.
\newblock Building text and speech datasets for low resourced languages: A case of languages in east africa.
\newblock In \emph{3rd Workshop on African Natural Language Processing}.

\bibitem[{Babu et~al.(2022)Babu, Wang, Tjandra, Lakhotia, Xu, Goyal, Singh, von Platen, Saraf, Pino, Baevski, Conneau, and Auli}]{Babu2022XLSRSC}
Arun Babu, Changhan Wang, Andros Tjandra, Kushal Lakhotia, Qiantong Xu, Naman Goyal, Kritika Singh, Patrick von Platen, Yatharth Saraf, Juan~Miguel Pino, Alexei Baevski, Alexis Conneau, and Michael Auli. 2022.
\newblock Xls-r: Self-supervised cross-lingual speech representation learning at scale.
\newblock In \emph{INTERSPEECH}.

\bibitem[{Baevski et~al.(2020{\natexlab{a}})Baevski, Schneider, and Auli}]{Baevski2020vqwav2vecSL}
Alexei Baevski, Steffen Schneider, and Michael Auli. 2020{\natexlab{a}}.
\newblock vq-wav2vec: Self-supervised learning of discrete speech representations.
\newblock \emph{ArXiv}, abs/1910.05453.

\bibitem[{Baevski et~al.(2020{\natexlab{b}})Baevski, Zhou, rahman Mohamed, and Auli}]{Baevski2020wav2vec2A}
Alexei Baevski, Henry Zhou, Abdel rahman Mohamed, and Michael Auli. 2020{\natexlab{b}}.
\newblock wav2vec 2.0: A framework for self-supervised learning of speech representations.
\newblock \emph{ArXiv}, abs/2006.11477.

\bibitem[{Baingana and Bos(2006)}]{baingana2006changing}
Florence~K Baingana and Eduard~R Bos. 2006.
\newblock Changing patterns of disease and mortality in sub-saharan africa: an overview.
\newblock \emph{Disease and Mortality in Sub-Saharan Africa. 2nd edition}.

\bibitem[{Blackley et~al.(2019)Blackley, Huynh, Wang, Korach, and Zhou}]{blackley2019speech}
Suzanne~V Blackley, Jessica Huynh, Liqin Wang, Zfania Korach, and Li~Zhou. 2019.
\newblock Speech recognition for clinical documentation from 1990 to 2018: a systematic review.
\newblock \emph{Journal of the american medical informatics association}, 26(4):324--338.

\bibitem[{Blackley et~al.(2020)Blackley, Schubert, Goss, Al~Assad, Garabedian, and Zhou}]{blackley2020physician}
Suzanne~V Blackley, Valerie~D Schubert, Foster~R Goss, Wasim Al~Assad, Pamela~M Garabedian, and Li~Zhou. 2020.
\newblock Physician use of speech recognition versus typing in clinical documentation: a controlled observational study.
\newblock \emph{International Journal of Medical Informatics}, 141:104178.

\bibitem[{Brown et~al.(2020)Brown, Mann, Ryder, Subbiah, Kaplan, Dhariwal, Neelakantan, Shyam, Sastry, Askell et~al.}]{brown2020language}
Tom Brown, Benjamin Mann, Nick Ryder, Melanie Subbiah, Jared~D Kaplan, Prafulla Dhariwal, Arvind Neelakantan, Pranav Shyam, Girish Sastry, Amanda Askell, et~al. 2020.
\newblock Language models are few-shot learners.
\newblock \emph{Advances in neural information processing systems}, 33:1877--1901.

\bibitem[{Bukachi and Pakenham-Walsh(2007)}]{bukachi2007information}
Frederick Bukachi and Neil Pakenham-Walsh. 2007.
\newblock Information technology for health in developing countries.
\newblock \emph{Chest}, 132(5):1624--1630.

\bibitem[{Chen et~al.(2021)Chen, Chai, Wang, Du, Zhang, Weng, Su, Povey, Trmal, Zhang, Jin, Khudanpur, Watanabe, Zhao, Zou, Li, Yao, Wang, Wang, You, and Yan}]{Chen2021GigaSpeechAE}
Guoguo Chen, Shuzhou Chai, Guan-Bo Wang, Jiayu Du, Weiqiang Zhang, Chao Weng, Dan Su, Daniel Povey, Jan Trmal, Junbo Zhang, Mingjie Jin, Sanjeev Khudanpur, Shinji Watanabe, Shuaijiang Zhao, Wei Zou, Xiangang Li, Xuchen Yao, Yongqing Wang, Yujun Wang, Zhao You, and Zhiyong Yan. 2021.
\newblock Gigaspeech: An evolving, multi-domain asr corpus with 10, 000 hours of transcribed audio.
\newblock In \emph{Interspeech}.

\bibitem[{Chen et~al.(2022)Chen, Wang, Chen, Wu, Liu, Chen, Li, Kanda, Yoshioka, Xiao et~al.}]{chen2022wavlm}
Sanyuan Chen, Chengyi Wang, Zhengyang Chen, Yu~Wu, Shujie Liu, Zhuo Chen, Jinyu Li, Naoyuki Kanda, Takuya Yoshioka, Xiong Xiao, et~al. 2022.
\newblock Wavlm: Large-scale self-supervised pre-training for full stack speech processing.
\newblock \emph{IEEE Journal of Selected Topics in Signal Processing}, 16(6):1505--1518.

\bibitem[{Chen et~al.(2020)Chen, Yang, feng Yeh, Jain, and Seltzer}]{Chen2020AipnetGA}
Yi-Chen Chen, Zhaojun Yang, Ching feng Yeh, Mahaveer Jain, and Michael~L. Seltzer. 2020.
\newblock Aipnet: Generative adversarial pre-training of accent-invariant networks for end-to-end speech recognition.
\newblock \emph{ICASSP 2020 - 2020 IEEE International Conference on Acoustics, Speech and Signal Processing (ICASSP)}, pages 6979--6983.

\bibitem[{Conneau et~al.(2021)Conneau, Baevski, Collobert, rahman Mohamed, and Auli}]{Conneau2021UnsupervisedCR}
Alexis Conneau, Alexei Baevski, Ronan Collobert, Abdel rahman Mohamed, and Michael Auli. 2021.
\newblock Unsupervised cross-lingual representation learning for speech recognition.
\newblock In \emph{Interspeech}.

\bibitem[{Das et~al.(2021)Das, Bodapati, Sunkara, Srinivasan, and Chau}]{Das2021BestOB}
Nilaksh Das, S.~Bodapati, Monica Sunkara, Sundararajan Srinivasan, and Duen~Horng Chau. 2021.
\newblock Best of both worlds: Robust accented speech recognition with adversarial transfer learning.
\newblock In \emph{Interspeech}.

\bibitem[{Davody et~al.(2022)Davody, Adelani, Kleinbauer, and Klakow}]{DBLP:conf/tsd/DavodyAKK22}
Ali Davody, David~Ifeoluwa Adelani, Thomas Kleinbauer, and Dietrich Klakow. 2022.
\newblock \href {https://doi.org/10.1007/978-3-031-16270-1\_12} {{TOKEN} is a {MASK:} few-shot named entity recognition with pre-trained language models}.
\newblock In \emph{Text, Speech, and Dialogue - 25th International Conference, {TSD} 2022, Brno, Czech Republic, September 6-9, 2022, Proceedings}, volume 13502 of \emph{Lecture Notes in Computer Science}, pages 138--150. Springer.

\bibitem[{De~Wet et~al.(2007)De~Wet, Louw, and Niesler}]{de2007human}
Febe De~Wet, Philippa Louw, and Thomas Niesler. 2007.
\newblock Human and automatic accent identification of nguni and sotho black south african english.
\newblock \emph{South African Journal of Science}, 103(3):159--164.

\bibitem[{Do{\u{g}}an et~al.(2014)Do{\u{g}}an, Leaman, and Lu}]{dougan2014ncbi}
Rezarta~Islamaj Do{\u{g}}an, Robert Leaman, and Zhiyong Lu. 2014.
\newblock Ncbi disease corpus: a resource for disease name recognition and concept normalization.
\newblock \emph{Journal of biomedical informatics}, 47:1--10.

\bibitem[{Dossou and Emezue(2021)}]{dossou2021okwugb}
Bonaventure~FP Dossou and Chris~C Emezue. 2021.
\newblock Okwugb$\backslash$'e: End-to-end speech recognition for fon and igbo.
\newblock \emph{arXiv preprint arXiv:2103.07762}.

\bibitem[{Doumbouya et~al.(2021)Doumbouya, Einstein, and Piech}]{doumbouya2021using}
Moussa Doumbouya, Lisa Einstein, and Chris Piech. 2021.
\newblock Using radio archives for low-resource speech recognition: towards an intelligent virtual assistant for illiterate users.
\newblock In \emph{Proceedings of the AAAI Conference on Artificial Intelligence}, volume~35, pages 14757--14765.

\bibitem[{Eberhard et~al.(2019)Eberhard, Simons, and Fennig}]{Ethnologue_Eberhard}
David Eberhard, Gary Simons, and Chuck Fennig. 2019.
\newblock \emph{Ethnologue: Languages of the World, 22nd Edition}.

\bibitem[{Etori et~al.(2023)Etori, Temesgen, and Gini}]{etori2023we}
Naome Etori, Ebasa Temesgen, and Maria Gini. 2023.
\newblock What we know so far: Artificial intelligence in african healthcare.
\newblock \emph{arXiv preprint arXiv:2305.18302}.

\bibitem[{Goss et~al.(2019)Goss, Blackley, Ortega, Kowalski, Landman, Lin, Meteer, Bakes, Gradwohl, Bates et~al.}]{goss2019clinician}
Foster~R Goss, Suzanne~V Blackley, Carlos~A Ortega, Leigh~T Kowalski, Adam~B Landman, Chen-Tan Lin, Marie Meteer, Samantha Bakes, Stephen~C Gradwohl, David~W Bates, et~al. 2019.
\newblock A clinician survey of using speech recognition for clinical documentation in the electronic health record.
\newblock \emph{International journal of medical informatics}, 130:103938.

\bibitem[{de~Graft~Aikins et~al.(2010)de~Graft~Aikins, Unwin, Agyemang, Allotey, Campbell, and Arhinful}]{de2010tackling}
Ama de~Graft~Aikins, Nigel Unwin, Charles Agyemang, Pascale Allotey, Catherine Campbell, and Daniel Arhinful. 2010.
\newblock Tackling africa's chronic disease burden: from the local to the global.
\newblock \emph{Globalization and health}, 6(1):1--7.

\bibitem[{Grosman(2021)}]{grosman2021xlsr53-large-english}
Jonatas Grosman. 2021.
\newblock Fine-tuned {XLSR}-53 large model for speech recognition in {E}nglish.
\newblock \url{https://huggingface.co/jonatasgrosman/wav2vec2-large-xlsr-53-english}.

\bibitem[{Gulati et~al.(2020)Gulati, Qin, Chiu, Parmar, Zhang, Yu, Han, Wang, Zhang, Wu et~al.}]{gulati2020conformer}
Anmol Gulati, James Qin, Chung-Cheng Chiu, Niki Parmar, Yu~Zhang, Jiahui Yu, Wei Han, Shibo Wang, Zhengdong Zhang, Yonghui Wu, et~al. 2020.
\newblock Conformer: Convolution-augmented transformer for speech recognition.
\newblock \emph{arXiv preprint arXiv:2005.08100}.

\bibitem[{Gutkin et~al.(2020)Gutkin, Demirsahin, Kjartansson, Rivera, and T{\'u}b{\`o}s{\'u}n}]{gutkin2020developing}
Alexander Gutkin, Isin Demirsahin, Oddur Kjartansson, Clara~E Rivera, and K{\'o}l{\'a} T{\'u}b{\`o}s{\'u}n. 2020.
\newblock Developing an open-source corpus of yoruba speech.

\bibitem[{Hassan et~al.(2022)Hassan, Rehmat, Ghani~Khan, Yousaf et~al.}]{hassan2022improvement}
Muhammad~Ahmed Hassan, Asim Rehmat, Muhammad~Usman Ghani~Khan, Muhammad~Haroon Yousaf, et~al. 2022.
\newblock Improvement in automatic speech recognition of south asian accent using transfer learning of deepspeech2.
\newblock \emph{Mathematical Problems in Engineering}, 2022.

\bibitem[{Heine and Nurse(2000)}]{heine2000african}
Bernd Heine and Derek Nurse. 2000.
\newblock \emph{African languages: An introduction}.
\newblock Cambridge University Press.

\bibitem[{Hernandez et~al.(2018)Hernandez, Nguyen, Ghannay, Tomashenko, and Esteve}]{hernandez2018ted}
Fran{\c{c}}ois Hernandez, Vincent Nguyen, Sahar Ghannay, Natalia Tomashenko, and Yannick Esteve. 2018.
\newblock Ted-lium 3: twice as much data and corpus repartition for experiments on speaker adaptation.
\newblock In \emph{International conference on speech and computer}, pages 198--208. Springer.

\bibitem[{Hsu et~al.(2021{\natexlab{a}})Hsu, Bolte, Tsai, Lakhotia, Salakhutdinov, and Mohamed}]{hsu2021hubert}
Wei-Ning Hsu, Benjamin Bolte, Yao-Hung~Hubert Tsai, Kushal Lakhotia, Ruslan Salakhutdinov, and Abdelrahman Mohamed. 2021{\natexlab{a}}.
\newblock Hubert: Self-supervised speech representation learning by masked prediction of hidden units.
\newblock \emph{IEEE/ACM Transactions on Audio, Speech, and Language Processing}, 29:3451--3460.

\bibitem[{Hsu et~al.(2021{\natexlab{b}})Hsu, Bolte, Tsai, Lakhotia, Salakhutdinov, and Mohamed}]{9585401}
Wei-Ning Hsu, Benjamin Bolte, Yao-Hung~Hubert Tsai, Kushal Lakhotia, Ruslan Salakhutdinov, and Abdelrahman Mohamed. 2021{\natexlab{b}}.
\newblock \href {https://doi.org/10.1109/TASLP.2021.3122291} {Hubert: Self-supervised speech representation learning by masked prediction of hidden units}.
\newblock \emph{IEEE/ACM Transactions on Audio, Speech, and Language Processing}, 29:3451--3460.

\bibitem[{Ibejih et~al.(2022)Ibejih, Oyewusi, Adekanmbi, and Osakuade}]{ibejih2022edustt}
Sharon Ibejih, Wuraola~Fisayo Oyewusi, Olubayo Adekanmbi, and Opeyemi Osakuade. 2022.
\newblock \href {https://openreview.net/forum?id=SCzGtqGVL-q} {{EDUSTT}: In-domain speech recognition for nigerian accented educational contents in english}.
\newblock In \emph{3rd Workshop on African Natural Language Processing}.

\bibitem[{Javed et~al.(2021)Javed, Doddapaneni, Raman, Bhogale, Ramesh, Kunchukuttan, Kumar, and Khapra}]{indicwav2vec}
Tahir Javed, Sumanth Doddapaneni, Abhigyan Raman, Kaushal~Santosh Bhogale, Gowtham Ramesh, Anoop Kunchukuttan, Pratyush Kumar, and Mitesh~M. Khapra. 2021.
\newblock \href {https://doi.org/10.48550/ARXIV.2111.03945} {Towards building asr systems for the next billion users}.

\bibitem[{Javed et~al.(2022)Javed, Doddapaneni, Raman, Bhogale, Ramesh, Kunchukuttan, Kumar, and Khapra}]{javed2022towards}
Tahir Javed, Sumanth Doddapaneni, Abhigyan Raman, Kaushal~Santosh Bhogale, Gowtham Ramesh, Anoop Kunchukuttan, Pratyush Kumar, and Mitesh~M Khapra. 2022.
\newblock Towards building asr systems for the next billion users.
\newblock In \emph{Proceedings of the AAAI Conference on Artificial Intelligence}, volume~36, pages 10813--10821.

\bibitem[{Kamper and Niesler(2011)}]{kamper2011multi}
Herman Kamper and Thomas Niesler. 2011.
\newblock Multi-accent speech recognition of afrikaans, black and white varieties of south african english.
\newblock In \emph{Twelfth Annual Conference of the International Speech Communication Association}.

\bibitem[{Kinfu et~al.(2009)Kinfu, Dal~Poz, Mercer, and Evans}]{kinfu2009health}
Yohannes Kinfu, Mario~R Dal~Poz, Hugo Mercer, and David~B Evans. 2009.
\newblock The health worker shortage in africa: are enough physicians and nurses being trained?

\bibitem[{Koenecke et~al.(2020)Koenecke, Nam, Lake, Nudell, Quartey, Mengesha, Toups, Rickford, Jurafsky, and Goel}]{koenecke2020racial}
Allison Koenecke, Andrew Nam, Emily Lake, Joe Nudell, Minnie Quartey, Zion Mengesha, Connor Toups, John~R Rickford, Dan Jurafsky, and Sharad Goel. 2020.
\newblock Racial disparities in automated speech recognition.
\newblock \emph{Proceedings of the National Academy of Sciences}, 117(14):7684--7689.

\bibitem[{Li et~al.(2021)Li, Manohar, Chitkara, Tjandra, Picheny, Zhang, Zhang, and Saraf}]{li2021accent}
Jialu Li, Vimal Manohar, Pooja Chitkara, Andros Tjandra, Michael Picheny, Frank Zhang, Xiaohui Zhang, and Yatharth Saraf. 2021.
\newblock Accent-robust automatic speech recognition using supervised and unsupervised wav2vec embeddings.
\newblock \emph{arXiv preprint arXiv:2110.03520}.

\bibitem[{Lodhi(1993)}]{lodhi1993language}
Abdulaziz~Y Lodhi. 1993.
\newblock The language situation in africa today.
\newblock \emph{Nordic Journal of African Studies}, 2(1):11--11.

\bibitem[{Loshchilov and Hutter(2017)}]{loshchilov2017decoupled}
Ilya Loshchilov and Frank Hutter. 2017.
\newblock Decoupled weight decay regularization.
\newblock \emph{arXiv preprint arXiv:1711.05101}.

\bibitem[{Manyati and Mutsau(2021)}]{manyati2021systematic}
Tarisai~Kudakwashe Manyati and Morgen Mutsau. 2021.
\newblock A systematic review of the factors that hinder the scale up of mobile health technologies in antenatal care programmes in sub-saharan africa.
\newblock \emph{African Journal of Science, Technology, Innovation and Development}, 13(1):125--131.

\bibitem[{Merity et~al.(2016)Merity, Xiong, Bradbury, and Socher}]{merity2016pointer}
Stephen Merity, Caiming Xiong, James Bradbury, and Richard Socher. 2016.
\newblock \href {http://arxiv.org/abs/1609.07843} {Pointer sentinel mixture models}.

\bibitem[{Naicker et~al.(2010)Naicker, Eastwood, Plange-Rhule, and Tutt}]{naicker2010shortage}
Saraladevi Naicker, John~B Eastwood, Jacob Plange-Rhule, and Roger~C Tutt. 2010.
\newblock Shortage of healthcare workers in sub-saharan africa: a nephrological perspective.
\newblock \emph{Clinical nephrology}, 74:S129--33.

\bibitem[{Naicker et~al.(2009)Naicker, Plange-Rhule, Tutt, and Eastwood}]{naicker2009shortage}
Saraladevi Naicker, Jacob Plange-Rhule, Roger~C Tutt, and John~B Eastwood. 2009.
\newblock Shortage of healthcare workers in developing countries--africa.
\newblock \emph{Ethnicity \& disease}, 19(1):60.

\bibitem[{Nkomazana et~al.(2015)Nkomazana, Mash, Shaibu, and Phaladze}]{nkomazana2015stakeholders}
Oathokwa Nkomazana, Robert Mash, Sheila Shaibu, and Nthabiseng Phaladze. 2015.
\newblock Stakeholders’ perceptions on shortage of healthcare workers in primary healthcare in botswana: focus group discussions.
\newblock \emph{PloS one}, 10(8):e0135846.

\bibitem[{Ogayo et~al.(2022)Ogayo, Neubig, and Black}]{ogayo2022building}
Perez Ogayo, Graham Neubig, and Alan~W Black. 2022.
\newblock Building african voices.
\newblock \emph{arXiv preprint arXiv:2207.00688}.

\bibitem[{Okagbue et~al.(2017)Okagbue, Opanuga, Adamu, Ugwoke, Obasi, and Eze}]{okagbue2017personal}
Hilary~I Okagbue, Abiodun~A Opanuga, Muminu~O Adamu, Paulinus~O Ugwoke, Emmanuela~CM Obasi, and Grace~A Eze. 2017.
\newblock Personal name in igbo culture: a dataset on randomly selected personal names and their statistical analysis.
\newblock \emph{Data in brief}, 15:72--80.

\bibitem[{Olaleye et~al.(2022)Olaleye, Oneaţă, and Kamper}]{Olaleye2022YFACCAY}
Kayode Olaleye, Dan Oneaţă, and Herman Kamper. 2022.
\newblock Yfacc: A yor{\`u}b{\'a} speech-image dataset for cross-lingual keyword localisation through visual grounding.
\newblock \emph{ArXiv}, abs/2210.04600.

\bibitem[{Oleribe et~al.(2019)Oleribe, Momoh, Uzochukwu, Mbofana, Adebiyi, Barbera, Williams, and Taylor-Robinson}]{oleribe2019identifying}
Obinna~O Oleribe, Jenny Momoh, Benjamin~SC Uzochukwu, Francisco Mbofana, Akin Adebiyi, Thomas Barbera, Roger Williams, and Simon~D Taylor-Robinson. 2019.
\newblock Identifying key challenges facing healthcare systems in africa and potential solutions.
\newblock \emph{International journal of general medicine}, 12:395.

\bibitem[{Panayotov et~al.(2015{\natexlab{a}})Panayotov, Chen, Povey, and Khudanpur}]{panayotov2015librispeech}
Vassil Panayotov, Guoguo Chen, Daniel Povey, and Sanjeev Khudanpur. 2015{\natexlab{a}}.
\newblock Librispeech: an asr corpus based on public domain audio books.
\newblock In \emph{2015 IEEE international conference on acoustics, speech and signal processing (ICASSP)}, pages 5206--5210. IEEE.

\bibitem[{Panayotov et~al.(2015{\natexlab{b}})Panayotov, Chen, Povey, and Khudanpur}]{7178964}
Vassil Panayotov, Guoguo Chen, Daniel Povey, and Sanjeev Khudanpur. 2015{\natexlab{b}}.
\newblock \href {https://doi.org/10.1109/ICASSP.2015.7178964} {Librispeech: An asr corpus based on public domain audio books}.
\newblock In \emph{2015 IEEE International Conference on Acoustics, Speech and Signal Processing (ICASSP)}, pages 5206--5210.

\bibitem[{Pawar and Shrawankar(2016)}]{pawar2016question}
Komal Pawar and Urmila Shrawankar. 2016.
\newblock Question systematization using templates.
\newblock In \emph{3rd International Conference on Computing for Sustainable Global Development}.

\bibitem[{Prasad and Jyothi(2020)}]{prasad-jyothi-2020-accents}
Archiki Prasad and Preethi Jyothi. 2020.
\newblock \href {https://www.aclweb.org/anthology/2020.acl-main.345} {How accents confound: Probing for accent information in end-to-end speech recognition systems}.
\newblock In \emph{Proceedings of the 58th Annual Meeting of the Association for Computational Linguistics}, pages 3739--3753, Online. Association for Computational Linguistics.

\bibitem[{Radford et~al.(2022)Radford, Kim, Xu, Brockman, McLeavey, and Sutskever}]{radford2022robust}
Alec Radford, Jong~Wook Kim, Tao Xu, Greg Brockman, Christine McLeavey, and Ilya Sutskever. 2022.
\newblock Robust speech recognition via large-scale weak supervision.
\newblock \emph{arXiv preprint arXiv:2212.04356}.

\bibitem[{Ravanelli et~al.(2021)Ravanelli, Parcollet, Plantinga, Rouhe, Cornell, Lugosch, Subakan, Dawalatabad, Heba, Zhong et~al.}]{ravanelli2021speechbrain}
Mirco Ravanelli, Titouan Parcollet, Peter Plantinga, Aku Rouhe, Samuele Cornell, Loren Lugosch, Cem Subakan, Nauman Dawalatabad, Abdelwahab Heba, Jianyuan Zhong, et~al. 2021.
\newblock Speechbrain: A general-purpose speech toolkit.
\newblock \emph{arXiv preprint arXiv:2106.04624}.

\bibitem[{Sanabria et~al.(2023)Sanabria, Bogoychev, Markl, Carmantini, Klejch, and Bell}]{sanabria2023edinburgh}
Ramon Sanabria, Nikolay Bogoychev, Nina Markl, Andrea Carmantini, Ondrej Klejch, and Peter Bell. 2023.
\newblock The edinburgh international accents of english corpus: Towards the democratization of english asr.
\newblock In \emph{ICASSP 2023-2023 IEEE International Conference on Acoustics, Speech and Signal Processing (ICASSP)}, pages 1--5. IEEE.

\bibitem[{Schneider et~al.(2019)Schneider, Baevski, Collobert, and Auli}]{schneider19_interspeech}
Steffen Schneider, Alexei Baevski, Ronan Collobert, and Michael Auli. 2019.
\newblock \href {https://doi.org/10.21437/Interspeech.2019-1873} {{wav2vec: Unsupervised Pre-Training for Speech Recognition}}.
\newblock In \emph{Proc. Interspeech 2019}, pages 3465--3469.

\bibitem[{Siminyu et~al.(2021)Siminyu, Kalipe, Orlic, Abbott, Marivate, Freshia, Sibal, Neupane, Adelani, Taylor et~al.}]{siminyu2021ai4d}
Kathleen Siminyu, Godson Kalipe, Davor Orlic, Jade Abbott, Vukosi Marivate, Sackey Freshia, Prateek Sibal, Bhanu Neupane, David~I Adelani, Amelia Taylor, et~al. 2021.
\newblock Ai4d--african language program.
\newblock \emph{arXiv preprint arXiv:2104.02516}.

\bibitem[{Sun et~al.(2018)Sun, feng Yeh, Hwang, Ostendorf, and Xie}]{Sun2018DomainAT}
Sining Sun, Ching feng Yeh, Mei-Yuh Hwang, Mari Ostendorf, and Lei Xie. 2018.
\newblock Domain adversarial training for accented speech recognition.
\newblock \emph{2018 IEEE International Conference on Acoustics, Speech and Signal Processing (ICASSP)}, pages 4854--4858.

\bibitem[{Sun et~al.(2017)Sun, Zhang, Xie, and Zhang}]{sun2017unsupervised}
Sining Sun, Binbin Zhang, Lei Xie, and Yanning Zhang. 2017.
\newblock An unsupervised deep domain adaptation approach for robust speech recognition.
\newblock \emph{Neurocomputing}, 257:79--87.

\bibitem[{Sunkara et~al.(2020)Sunkara, Ronanki, Dixit, Bodapati, and Kirchhoff}]{Sunkara2020}
Monica Sunkara, Srikanth Ronanki, Kalpit Dixit, Sravan Bodapati, and Katrin Kirchhoff. 2020.
\newblock \href {https://www.amazon.science/publications/robust-prediction-of-punctuation-and-truecasing-for-medical-asr} {Robust prediction of punctuation and truecasing for medical asr}.
\newblock In \emph{ACL 2020 Workshop on NLP for Medical Conversations}.

\bibitem[{Valk and Alum{\"a}e(2021)}]{Valk2021VOXLINGUA107AD}
J{\"o}rgen Valk and Tanel Alum{\"a}e. 2021.
\newblock Voxlingua107: A dataset for spoken language recognition.
\newblock \emph{2021 IEEE Spoken Language Technology Workshop (SLT)}, pages 652--658.

\bibitem[{Vogel et~al.(2015)Vogel, Kaisers, Wassmuth, Mayatepek et~al.}]{vogel2015analysis}
Markus Vogel, Wolfgang Kaisers, Ralf Wassmuth, Ertan Mayatepek, et~al. 2015.
\newblock Analysis of documentation speed using web-based medical speech recognition technology: randomized controlled trial.
\newblock \emph{Journal of medical Internet research}, 17(11):e5072.

\bibitem[{Wheeler et~al.(2007)Wheeler, Barrett, Benson, Bryant, Canese, Chetvernin, Church, DiCuccio, Edgar, Federhen et~al.}]{wheeler2007database}
David~L Wheeler, Tanya Barrett, Dennis~A Benson, Stephen~H Bryant, Kathi Canese, Vyacheslav Chetvernin, Deanna~M Church, Michael DiCuccio, Ron Edgar, Scott Federhen, et~al. 2007.
\newblock Database resources of the national center for biotechnology information.
\newblock \emph{Nucleic acids research}, 36(suppl\_1):D13--D21.

\bibitem[{{Wikipedia contributors}(2023{\natexlab{a}})}]{enwiki:1132870977}
{Wikipedia contributors}. 2023{\natexlab{a}}.
\newblock Demographics of africa --- {Wikipedia}{,} the free encyclopedia.
\newblock \url{https://en.wikipedia.org/w/index.php?title=Demographics_of_Africa&oldid=1132870977}.
\newblock [Online; accessed 20-January-2023].

\bibitem[{{Wikipedia contributors}(2023{\natexlab{b}})}]{enwiki:1133594141}
{Wikipedia contributors}. 2023{\natexlab{b}}.
\newblock Languages of africa --- {Wikipedia}{,} the free encyclopedia.
\newblock \url{https://en.wikipedia.org/w/index.php?title=Languages_of_Africa&oldid=1133594141}.
\newblock [Online; accessed 20-January-2023].

\bibitem[{{Wikipedia contributors}(2023{\natexlab{c}})}]{enwiki:1146587606}
{Wikipedia contributors}. 2023{\natexlab{c}}.
\newblock List of cities in africa by population --- {Wikipedia}{,} the free encyclopedia.
\newblock \url{https://en.wikipedia.org/w/index.php?title=List_of_cities_in_Africa_by_population&oldid=1146587606}.
\newblock [Online; accessed 31-March-2023].

\bibitem[{Yadavalli et~al.(2022)Yadavalli, Mirishkar, and Vuppala}]{Yadavalli2022MultiTaskEM}
Aditya Yadavalli, Ganesh~S Mirishkar, and Anil~Kumar Vuppala. 2022.
\newblock Multi-task end-to-end model for telugu dialect and speech recognition.
\newblock In \emph{Interspeech}.

\bibitem[{Yao et~al.(2022)Yao, Dong, Zhang, Zhang, Xie, Liu, Lin, Sun, and Wang}]{yao2022prompt}
Yuan Yao, Bowen Dong, Ao~Zhang, Zhengyan Zhang, Ruobing Xie, Zhiyuan Liu, Leyu Lin, Maosong Sun, and Jianyong Wang. 2022.
\newblock Prompt tuning for discriminative pre-trained language models.
\newblock \emph{arXiv preprint arXiv:2205.11166}.

\bibitem[{Zhang et~al.(2022)Zhang, Zhang, Halpern, Patel, and Scharenborg}]{zhang22n_interspeech}
Yuanyuan Zhang, Yixuan Zhang, Bence Halpern, Tanvina Patel, and Odette Scharenborg. 2022.
\newblock \href {https://doi.org/10.21437/Interspeech.2022-836} {{Mitigating bias against non-native accents}}.
\newblock In \emph{Proc. Interspeech 2022}, pages 3168--3172.

\end{thebibliography}
\bibliographystyle{acl_natbib}



\appendix

\section{Appendix}

\subsection{Transcript Preprocessing}\label{appendix:transcripts}

  \paragraph{Date and Time replacement:}  Dates are a critical part of clinical documentation as they typically contain several references to dates and times, for example, date of admission, date of discharge, time of death, and so on. Sampled subsets of sentences containing data and time references from the clinical and general domain were randomly replaced with random dates and times in different formats including "10/12/1999", "10th December, 1999", "10th Dec, 1999", "10-12-1999", "Mon 10 Dec, 1999", "Monday 10th December, 1999". Similar timestamp variations were added to our templates.
  \paragraph{Cleaning:}  Final corpus was pre-processed and cleaned by splitting on sentence boundaries, normalizing spaces, removing carriage return characters, removing non-alphanumeric characters except those with important structural or semantic meaning in the clinical domain such as question marks, parenthesis, colon, a hyphen, plus sign, and greater/lesser than sign. We removed Transcripts with less than 5 characters and greater than 300 characters.
  \paragraph{Privacy and Patient information:} Although the clinical corpora used were already anonymized, we re-examined several sentence samples for inadvertent exposure of patient names. Anonymized datasets with de-identification tokens like [NAME] and [DATE] were replaced with African names and randomly generated dates as described above.

\subsection{Annotation Instructions}\label{appendix:annotation}
Recorders were provided with the following instructions:

    \paragraph{Accuracy} It is very important that the recorded words match the text in the script exactly. If you accidentally deviate from the script, become unsure, or lose track of your thought, please delete and record the prompt again.
    \paragraph{Punctuations} All punctuations should be pronounced in full, not just observed. That is, when reading a text sample that contains punctuation, you say "comma", "full stop", "semi-colon", "colon", "slash", "hyphen", "question mark", "exclamation mark", and so on as appropriate. Brackets should be pronounced as "open bracket" or "close bracket".
    \paragraph{Punctuation Exclusions/Exceptions}  to the above rule: In measurements or units like "mg/dl", please say "milligram PER dl" NOT "milligram slash dl". In situations where "?" is used to represent "query", please say "query" NOT "question mark".
    \paragraph{Abbreviations} Pronounce common short-hand forms (such as r/o, prn, tds, PO, mg, W/O), dates, times, and numbers as you would in a clinical setting. For example, "r/o" should be pronounced "rule out" as usual not "arr slash ohh".  Common Abbreviations SHOULD be pronounced in full. "CT" should be pronounced "see tee" as usual NOT "Computed Tomography". "CXR" should be pronounced "Chest Xray" as usual NOT "see ex arr" "mmHg" should be pronounced in full as "millimeters of mercury". Pronounce CA as "Carcinoma" NOT "See Ay".
    \paragraph{Tone} Also be sure to use your natural accent. The goal is to build a speech-to-text system that understands African accents. This tool is for us. Be natural.
    \paragraph{Speed} Do not speak unrealistically fast. While an increased reading speed is recommended, take care to avoid vocal fatigue from rushing through the phrases at lightning speed! This will only result in a lower-quality voice. Record a maximum of 2 hours a day, taking a break every half hour.
    
\subsection{Annotator Management}

\paragraph{Consent} Recorders signed a Terms of Use agreement and consented to the privacy policy on the recording platform.

\paragraph{Payment}Recorders were paid \$5 to \$10 per hour depending on task difficulty and clinical experience. Most recorders considered payment satisfactory compared with task difficulty.

\subsection{AfriSpeech Vocabulary}\label{appendix:vocabulary}
AfriSpeech models use a 50-character vocab including numbers and punctuations and symbols with important semantic roles in healthcare.

"-", "w", "a", "7", ",", "0", "d", "i", ":", "p", "g", "u", "(", "5", "1", "e", "9", "j", "b", "3", "s", "'", "h", "o", "+", "l", "v", "y", "q", "n", "2", "r", "f", "m", "\%", "t", "/", "6", "z", "?", "8", ")", "x", ".", "4", "c", "k", "|", "[UNK]", "[PAD]"

\end{document}